\pdfoutput=1
\PassOptionsToPackage{table,xcdraw,svgnames,dvipsnames}{xcolor} 

\documentclass[11pt]{article}

\usepackage[final]{acl}
\usepackage{times}
\usepackage{latexsym}
\usepackage{amsmath}
\usepackage{amssymb}
\usepackage[T1]{fontenc}

\usepackage[utf8]{inputenc}

\usepackage{microtype}

\usepackage{inconsolata}

\usepackage{graphicx}

\usepackage{algorithm}
\usepackage{algorithmic}
\usepackage{graphicx}               
\usepackage{booktabs}               
\usepackage{multirow}               
\usepackage{makecell}               
\usepackage[table]{xcolor}          

\title{Inefficiencies of Meta Agents for Agent Design}

\author{
 \textbf{Batu El \textsuperscript{$\pi$}}
 \textbf{Mert Yuksekgonul \textsuperscript{$\pi$}}
 \textbf{James Zou \textsuperscript{$\pi$}}
\\
\\
 \textsuperscript{$\pi$}Stanford University
\\
 \small{
   {\{batuel, merty, jamesz\}@stanford.edu}
 }
}

\begin{document}
\maketitle
\begin{abstract}
Recent works began to automate the design of agentic systems using meta-agents that propose and iteratively refine new agent architectures. In this paper, we examine three key challenges in a common class of meta-agents. \emph{First}, we investigate how a meta-agent learns across iterations and find that simply expanding the context with all previous agents, as proposed by previous works, performs worse than ignoring prior designs entirely. We show that the performance improves with an evolutionary approach. \emph{Second}, although the meta-agent designs multiple agents during training, it typically commits to a single agent at test time. We find that the designed agents have low behavioral diversity, limiting the potential for their complementary use. \emph{Third}, we assess when automated design is economically viable. We find that only in a few cases—specifically, two datasets—the overall cost of designing and deploying the agents is lower than that of human-designed agents when deployed on over 15,000 examples. In contrast, the performance gains for other datasets do not justify the design cost, regardless of scale. \href{https://github.com/batu-el/meta-agent-inefficiency}{\raisebox{-0.1\height}{\includegraphics[height=1em]
{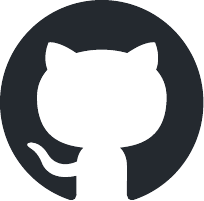}}}
\end{abstract}

\section{Introduction}
\begin{figure*}[t]
  \centering
  \includegraphics[width=\textwidth]{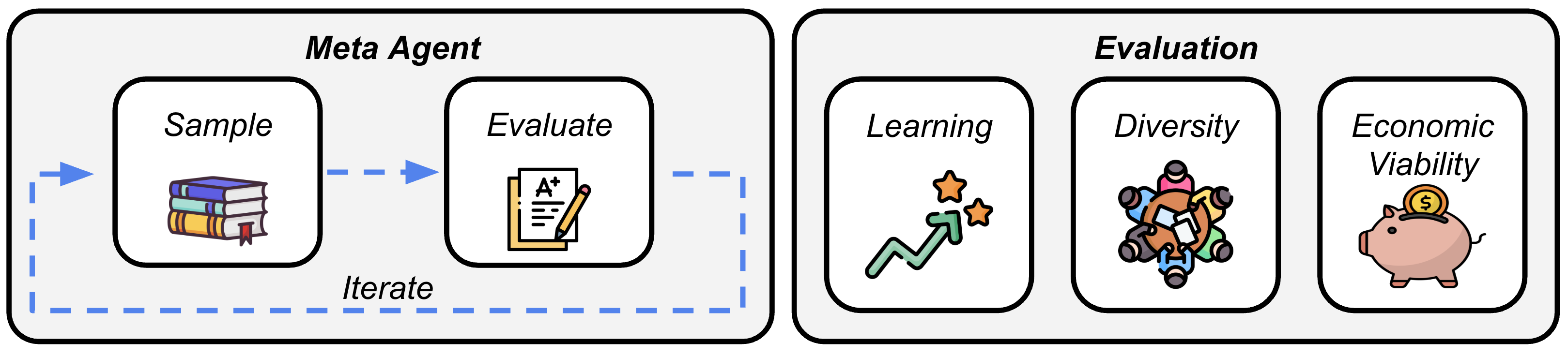}
  \caption{\textbf{Overview of the meta-agent framework.} The Meta-Agent iteratively
  samples and evaluates agents, refining its outputs through a feedback loop. We focus on three key dimensions:
  (1) learning from previously designed agents;
  (2) diversity and complementarity of generated agents; and
  (3) economic viability.}
  \label{fig:overview}
\end{figure*}

Agentic systems powered by language models demonstrated remarkable abilities to perform complex tasks and became a transformative force in many domains, including cutting-edge research and development  \citep{Swanson2024, lu2024aiscientistfullyautomated, yamada2025aiscientistv2workshoplevelautomated}, financial services \citep{okpala2025agenticaisystemsapplied, xiao2025tradingagentsmultiagentsllmfinancial}, and task automation \citep{fourney2024magenticonegeneralistmultiagentsolving}. Until recently, these systems were designed by researchers who built their domain knowledge into their agent architectures. However, a persistent trend in machine learning research, known as the Bitter Lesson \citep{sutton2019bitter}, suggests that hand-designed solutions are eventually replaced by solutions developed via scalable approaches that leverage \textit{search} and \textit{learning}. To this end, recent works have taken the first steps in the direction of automating the design of agentic systems 
\citep{hu2024automateddesignagenticsystems, li2024autoflowautomatedworkflowgeneration, saadfalcon2024archonarchitecturesearchframework, niu2025flowmodularizedagenticworkflow, nie2025weakforstrongtrainingweakmetaagent, shang2025agentsquareautomaticllmagent, wang2025scoreflowmasteringllmagent, ye2025masgpttrainingllmsbuild, zhang2025aflowautomatingagenticworkflow, zhang2025multiagentarchitecturesearchagentic}. Our work focuses on a common class of meta-agents that follow the \emph{sample–evaluate–iterate} pattern (see Figure~\ref{fig:overview}, Algorithm \ref{alg:meta-agent}) and highlights three challenges.

\paragraph{Meta Learning} We begin by examining the assumption that the meta-agent effectively learns from previously discovered agents. Our analysis reveals that the meta-agent framework proposed by \citet{hu2024automateddesignagenticsystems} does not meaningfully leverage prior designs. In fact, it performs worse than a baseline that ignores prior designs entirely. In contrast, we demonstrate that an evolutionary context curation strategy, where the generation of the next agent is conditioned on the previous  best-performing agents (parents), yields improved performance.
\paragraph{Diversity and Complementarity} While the meta-agent generates a set of candidate agents, typically only one is deployed, neglecting potential synergies among them. If the designed agents were behaviorally diverse, where each specializes in particular types of queries, this would enable dynamic selection of the most suitable agent per query. However, we find that the designed agents often lack behavioral diversity, which is even more pronounced when evolutionary strategies are used.

\paragraph{Economic Viability} For a meta-agent to be economically viable, the fixed cost of designing a new agent must be justified by corresponding improvements in performance. We formalize this trade-off by defining the total cost of a meta-designed agent as the sum of a fixed design cost and a per-example inference cost. This raises the key question: \textit{How many test examples are needed before the cost per correct response becomes lower when using the designed agent?} In our experiments, we find this break-even point occurs at approximately $15{,}000$ examples for MMLU and DROP. In contrast, for other datasets, the performance gains do not justify the design cost, regardless of the scale of deployment.

\section{Related Works}

\begin{table*}[t]
  \centering
  \footnotesize
  \setlength{\tabcolsep}{4pt}
  \resizebox{\textwidth}{!}{%
  \begin{tabular}{lcccccccccccccccc}
    \toprule
    \multirow{2}{*}{Dataset} &
      \multicolumn{3}{c}{Best Agent} &
      \multicolumn{3}{c}{Best-5 Avg.} &
      \multicolumn{3}{c}{Best-10 Avg.} &
      \multicolumn{3}{c}{Best-15 Avg.} &
      \multicolumn{4}{c}{Test Performance (Best Agent)} \\
    \cmidrule(lr){2-4}\cmidrule(lr){5-7}\cmidrule(lr){8-10}\cmidrule(lr){11-13}\cmidrule(lr){14-17}
      & C & P & E & C & P & E & C & P & E & C & P & E & I & C & P & E \\
    \midrule
    DROP &
      \shortstack{71.4\\(2.0)} &
      \shortstack{72.5\\(4.2)} &
      \multicolumn{1}{>{\columncolor{blue!20}}c}{\shortstack{\textbf{74.4}\\(3.2)}} &
      \shortstack{68.1\\(1.1)} &
      \shortstack{69.3\\(1.2)} &
      \multicolumn{1}{>{\columncolor{blue!20}}c}{\shortstack{\textbf{71.5}\\(4.5)}} &
      \shortstack{66.6\\(0.8)} &
      \shortstack{66.8\\(1.1)} &
      \multicolumn{1}{>{\columncolor{blue!20}}c}{\shortstack{\textbf{69.7}\\(4.7)}} &
      \shortstack{64.9\\(0.6)} &
      \shortstack{64.9\\(1.8)} &
      \multicolumn{1}{>{\columncolor{blue!20}}c}{\shortstack{\textbf{68.2}\\(4.7)}} &
      \shortstack{64.8\\(1.3)} &
      \shortstack{71.9\\(3.2)} &
      \shortstack{72.6\\(7.8)} &
      \multicolumn{1}{>{\columncolor{blue!20}}c}{\shortstack{\textbf{73.2}\\(5.1)}} \\[1pt]

    MGSM &
      \shortstack{41.4\\(6.2)} &
      \shortstack{56.2\\(10.5)} &
      \multicolumn{1}{>{\columncolor{blue!20}}c}{\shortstack{\textbf{56.5}\\(4.7)}} &
      \shortstack{32.5\\(13.8)} &
      \shortstack{48.4\\(9.8)} &
      \multicolumn{1}{>{\columncolor{blue!20}}c}{\shortstack{\textbf{50.4}\\(0.8)}} &
      \shortstack{27.4\\(16.6)} &
      \shortstack{43.4\\(7.9)} &
      \multicolumn{1}{>{\columncolor{blue!20}}c}{\shortstack{\textbf{46.0}\\(1.6)}} &
      \shortstack{22.4\\(17.1)} &
      \shortstack{39.8\\(5.3)} &
      \multicolumn{1}{>{\columncolor{blue!20}}c}{\shortstack{\textbf{42.7}\\(2.5)}} &
      \shortstack{38.4\\(2.8)} &
      \shortstack{41.2\\(4.8)} &
      \shortstack{51.8\\(7.6)} &
      \multicolumn{1}{>{\columncolor{blue!20}}c}{\shortstack{\textbf{53.5}\\(2.0)}} \\[1pt]

    MMLU &
      \shortstack{74.7\\(2.0)} &
      \shortstack{76.3\\(1.6)} &
      \multicolumn{1}{>{\columncolor{blue!20}}c}{\shortstack{\textbf{76.6}\\(2.7)}} &
      \shortstack{73.0\\(2.1)} &
      \shortstack{73.8\\(2.4)} &
      \multicolumn{1}{>{\columncolor{blue!20}}c}{\shortstack{\textbf{74.8}\\(2.7)}} &
      \shortstack{70.3\\(5.2)} &
      \shortstack{72.4\\(2.5)} &
      \multicolumn{1}{>{\columncolor{blue!20}}c}{\shortstack{\textbf{73.7}\\(2.4)}} &
      \shortstack{68.0\\(8.0)} &
      \shortstack{71.1\\(3.0)} &
      \multicolumn{1}{>{\columncolor{blue!20}}c}{\shortstack{\textbf{72.7}\\(2.3)}} &
      \shortstack{62.8\\(2.3)} &
      \shortstack{66.2\\(4.2)} &
      \multicolumn{1}{>{\columncolor{blue!20}}c}{\shortstack{\textbf{67.8}\\(0.8)}} &
      \shortstack{65.8\\(3.3)} \\[1pt]

    GPQA &
      \shortstack{32.3\\(2.6)} &
      \multicolumn{1}{>{\columncolor{blue!20}}c}{\shortstack{\textbf{35.2}\\(2.8)}} &
      \shortstack{33.8\\(2.2)} &
      \shortstack{26.4\\(8.7)} &
      \multicolumn{1}{>{\columncolor{blue!20}}c}{\shortstack{\textbf{32.2}\\(1.5)}} &
      \shortstack{31.2\\(0.9)} &
      \shortstack{22.5\\(12.4)} &
      \multicolumn{1}{>{\columncolor{blue!20}}c}{\shortstack{\textbf{30.4}\\(1.1)}} &
      \shortstack{29.8\\(0.8)} &
      \shortstack{20.7\\(13.2)} &
      \multicolumn{1}{>{\columncolor{blue!20}}c}{\shortstack{\textbf{29.1}\\(0.6)}} &
      \shortstack{28.8\\(0.5)} &
      \shortstack{30.0\\(2.4)} &
      \shortstack{29.7\\(2.7)} &
      \multicolumn{1}{>{\columncolor{blue!20}}c}{\shortstack{\textbf{31.3}\\(0.0)}} &
      \shortstack{28.5\\(3.1)} \\
    \midrule
     \midrule
Avg. &
  \shortstack{55.0} &
  \multicolumn{1}{>{\columncolor{green!20}}c}{\shortstack{\textbf{60.0}}} &
  \multicolumn{1}{>{\columncolor{green!20}}c}{\shortstack{\textbf{60.3}}} &
  \shortstack{50.0} &
  \multicolumn{1}{>{\columncolor{green!20}}c}{\shortstack{\textbf{55.9}}} &  \multicolumn{1}{>{\columncolor{green!20}}c}{\shortstack{\textbf{57.0}}} &
  \shortstack{46.7} &
  \multicolumn{1}{>{\columncolor{green!20}}c}{\shortstack{\textbf{53.2}}} &  \multicolumn{1}{>{\columncolor{green!20}}c}{\shortstack{\textbf{54.8}}} &
  \shortstack{44.0} &
  \multicolumn{1}{>{\columncolor{green!20}}c}{\shortstack{\textbf{51.2}}} &
  \multicolumn{1}{>{\columncolor{green!20}}c}{\shortstack{\textbf{53.1}}} &
  \shortstack{49.0} &
  \shortstack{52.2} &
  \multicolumn{1}{>{\columncolor{green!20}}c}{\shortstack{\textbf{55.9}}} &
  \multicolumn{1}{>{\columncolor{green!20}}c}{\shortstack{\textbf{55.2}}} \\
    \bottomrule
  \end{tabular}}
\caption{\textbf{Meta-Agent Performance: Parallel context curation outperforms cumulative curation, while evolutionary approaches lead to further improvements.} Columns $1$–$12$ report performance on $D_{\text{train}}$ for: the single \emph{Best Agent} (cols $1$–$3$), and the averages of the top 5 (cols $4$–$6$), top 10 (cols $7$–$9$), and top 15 (cols $10$–$12$) agents, evaluated under three context curation strategies: Cumulative (C), Parallel (P), and Evolutionary (E).  
Columns $13$–$16$ show the $D_{\text{test}}$ performance of the agent that achieves the highest score on $D_{\text{train}}$. I denotes the test performance of the best agent from the Initial library selected based on its training performance. Averaged across $3$ runs.}

  \label{tab:full_results}
\end{table*}

Our primary reference is ADAS \citep{hu2024automateddesignagenticsystems}, which has introduced meta-agent search with the idea of searching for agents in the code space. MAS-GPT \citep{ye2025masgpttrainingllmsbuild} and ScoreFlow \citep{wang2025scoreflowmasteringllmagent} develop meta-agents by training a model to dynamically generate multi-agent systems for a given query. AgentSquare \citep{shang2025agentsquareautomaticllmagent} and Archon \citep{saadfalcon2024archonarchitecturesearchframework} explore modular agent architectures and use discrete module recombination to efficiently search design spaces. AutoFlow \citep{li2024autoflowautomatedworkflowgeneration}, Weak-for-Strong \citep{nie2025weakforstrongtrainingweakmetaagent}, and ADAS \citep{hu2024automateddesignagenticsystems} use a meta agent that follows the sample-evaluate-iterate paradigm (Algorithm \ref{alg:meta-agent}).
Other recent meta-agent approaches include Multi-agent Supernet \citep{zhang2025multiagentarchitecturesearchagentic}, Flow \citep{niu2025flowmodularizedagenticworkflow}, and AFlow \citep{zhang2025aflowautomatingagenticworkflow}. \citet{erol2025costofpasseconomicframeworkevaluating} examined the cost of producing a correct response, which is directly relevant to our economic viability analysis.

\begin{algorithm}
\caption{Meta Agent: \textit{Sample-Evaluate-Iterate}}
\label{alg:meta-agent}
\begin{algorithmic}[1]
\STATE $D_{\text{train}}$ \textcolor{gray}{\# set of training examples}
\STATE $F$ \textcolor{gray}{\# initial agents library}
\STATE $A = \{(f_{0_i}, s_{0_i}) \mid f_{0_i} \in F\}$ \textcolor{gray}{\# archive}
\FOR{$t$ in $[T]$} 
    \STATE $\hat{A} = \phi(A)$ \textcolor{gray}{\# select current context}
    \STATE $f_t \sim \Pi(\cdot \mid \hat{A})$ \textcolor{blue}{\# sample}\textcolor{gray}{, revise, debug}
     \STATE $s_{t} = eval(f_t)$ \textcolor{blue}{\# evaluate}
    \STATE $A.{append}(f_t, s_{t})$ \textcolor{gray}{\# add to archive}
    \STATE \textcolor{blue}{\# iterate}
\ENDFOR
\end{algorithmic}
\end{algorithm}
\section{Setup}
Following \citet{hu2024automateddesignagenticsystems}, we define an agent as a computer program that takes a question as input and makes language model calls to compute the answer. Let \(f_i\) denote an agent and score \(s_i = {eval}(f_i, D_{\text{train}}) \in \mathbb{R}^{N_{\text{train}}}\) be the evaluation vector containing the agent $i$'s evaluation scores for each example in the training dataset $D_{\text{train}}$. The agent $f_i$ is represented by code. The archive, $A$, is a set of discovered agents \(\{f_i\}\) and their corresponding evaluations on the training set. We initialize the archive with the agents in the initial agents library, $F$, and their corresponding evaluations.\footnote{The content of the initial agents library is discussed in Appendix \ref{apdx:initial-agents}.} At each iteration, the meta-agent samples a new agent design using a language model, $\Pi$, conditioned on a curated subset of the current archive, $\hat{A}$. The function \(\phi\) implements this curation step. The sampling step is followed by revisions to ensure proper formatting and debugging with execution feedback. Finally, the new agent, $f_t$, is added to the archive $A$.\footnote{Appendix \ref{apdx:exp-setup} details the configurations we use in our experiments.} Algorithm \ref{alg:meta-agent} outlines the design procedure. We experiment with three instantiations of context curation~(\(\phi\)):

\paragraph{Cumulative.} \(\phi_{C}\) is identity, and the generation of the next agent is conditioned on all the previously discovered architectures, as in \citet{hu2024automateddesignagenticsystems}.
\paragraph{Parallel.} \(\phi_{P}\) maps any archive to only the subset that contains $7$ agents in the initial library and corresponding evaluation scores. Hence, the meta-agent ignores the previously designed architectures, effectively parallel sampling the new agents.

\paragraph{Evolutionary.} \(\phi_{E}\) selects a subset of $7$ agents with the best evaluation scores (\textit{top-k} selection strategy) from $A$ to be the \textit{parents} of the next agent generation. The generation of the next agent is conditioned on this higher-quality subset of the previously discovered architectures at each iteration. 

\paragraph{Tasks and Models} 
Closely following the prior work~\citep{hu2024automateddesignagenticsystems}, we evaluate our agentic design setup on 1) mathematical reasoning abilities in a multi-lingual setting, MGSM, \citep{shi2022languagemodelsmultilingualchainofthought}, 2) reading comprehension, DROP, \citep{dua-etal-2019-drop}, 3) multi-task problem solving, MMLU, \citep{hendrycks2021measuringmassivemultitasklanguage}, and 4) graduate-level science questions, GPQA \citep{rein2023gpqagraduatelevelgoogleproofqa}. From these datasets, we sample disjoint subsets \( D_{\text{train}} \) to compute $s_i$, and \( D_{\text{test}} \) to be used as held-out evaluation. The details of our experimental setup are explained in Appendix \ref{apdx:exp-setup}. All the results we report are averaged across $3$ runs.

\section{Experiments}
\subsection{Learning}
Table~\ref{tab:full_results} compares three context curation strategies for meta-agent design. We find that \emph{cumulative context curation does not outperform parallel context curation}, suggesting that ADAS-style meta-agents derive limited benefits from prior agent designs and perform worse than ignoring prior designs entirely. 

In contrast, \emph{evolutionary context curation improves performance}, yielding up to a $+10\%$ gain over cumulative context on MGSM. This suggests that selectively including high-quality prior designs in context enables more effective meta-learning.


\subsection{Diversity and Complementarity}

\begin{figure*}[t]
  \centering
  \includegraphics[width=\textwidth]{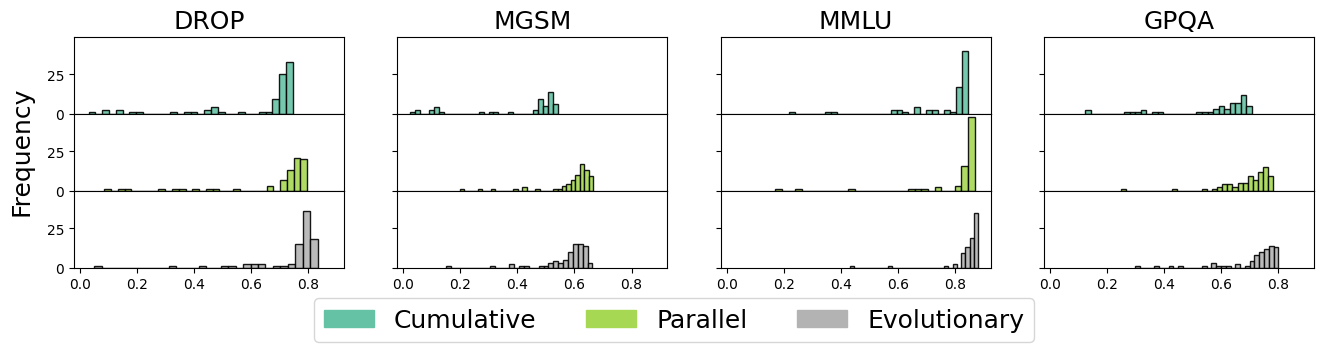}
\caption{\textbf{Agent Diversity: Cumulative context curation yields lower overall similarity. Parallel context curation produces greater agent diversity compared to evolutionary curation, highlighting an exploration exploitation trade-off.} Histograms of agent similarities (row averages of $\mathbf{C}$), excluding agents with zero performance (all-black rows of $\mathbf{S}$ in Figure \ref{fig:S}, and corresponding dark blue rows and columns of $\mathbf{C}$ in Figure \ref{fig:SST}). Each subplot shows histograms of averaged similarity scores for each agent (x-axis) and their frequency (y-axis) across $3$ runs.}
  \label{fig:Figure6_rowavg}
\end{figure*}
To investigate the potential synergies between the generated agents, we turn our attention to the behavioral diversity of the agent pool and analyze whether the agents have similar behavior on training examples. \textit{How often the questions they get right overlap? Do they make the same mistakes?}

We analyze agent diversity by computing similarities between evaluation vectors. Let $s_i = \text{eval}(f_i, D_{\text{train}}) \in \mathbb{R}^{N_{\text{train}}}$ be the evaluation vector for agent $f_i$. Stacking $s_i$ as rows, we obtain $\mathbf{S}$, which, in effect, represents embeddings of each agent from the perspective of the training questions (see Figure \ref{fig:S}). We then compute the cosine similarity matrix \(\mathbf{C} \), where the entry $i,j$ corresponds to the cosine similarity $\langle s_i, s_j\rangle$ (see Figure \ref{fig:SST}). This pairwise similarity metric favors agents that succeed on the same examples. We show the histograms of pairwise similarities (entries of  \(\mathbf{C}\)) in Figure \ref{fig:Figure6_entries} and the histograms for the average similarity of an agent to the rest of the agents (row averages of  \(\mathbf{C}\)) in Figure \ref{fig:Figure6_rowavg}.

Figure \ref{fig:Figure6_rowavg} shows the  similarity distributions, with evolutionary context curation generally exhibiting higher similarity scores. We observe that \emph{cumulative context curation yield lower similarity} overall compared to parallel and evolutionary context curation. Moreover, while \emph{parallel and evolutionary context curation yield similar performance}, parallel context curation exhibits slightly lower similarity and produces more diverse agents. Notably, in GPQA, parallel context curation yields both better-performing and more diverse agents. 

Our analysis of coverage (Table~\ref{tab:coverage})—the proportion of questions correctly answered at least by one of the designed agents—shows that \emph{parallel context curation yields the highest coverage}, highlighting its effectiveness in promoting exploration.
\subsection{Economic Viability}
\begin{figure}[h]
  \centering
\includegraphics[width=0.45\textwidth]{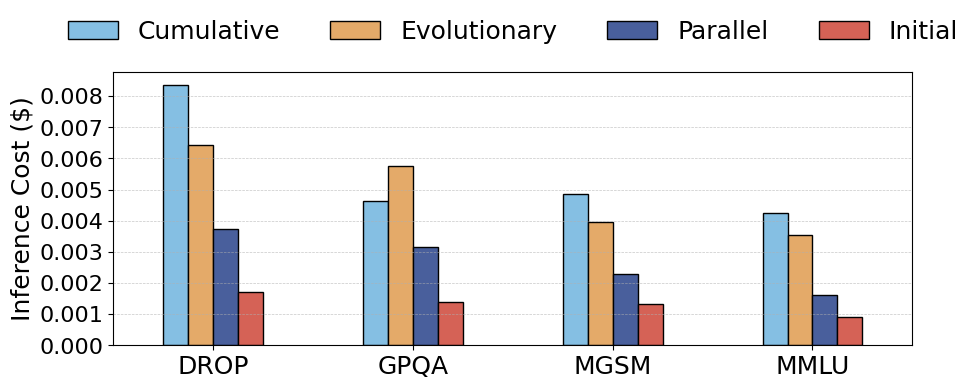}
  \caption{\textbf{Average inference cost per test query: C > E > P > I. } For agents in the initial library $F$ (Initial, see Appendix \ref{apdx:initial-agents}), agents designed by meta agent with \(\phi_C\) (Cumulative), agents designed by meta agent with \(\phi_P\) (Parallel) , agents designed by meta agent with \(\phi_E\) (Evolutionary). Averaged across all agents from 3 runs.}
  \label{fig:fig3_1}
\end{figure}

In Figure \ref{fig:fig3_1}, we observe that the agents designed using \(\phi_C\) have the highest average inference costs, followed by those designed using evolutionary context, \(\phi_E\). Among the meta agents, the one that uses parallel context curation produces the least costly agents on average,  a trend also observed among the best-performing agents (Figure \ref{fig:Figure_BestAgentCost}). However, agents designed by the meta agent still remain more costly than those in the initial library.

To identify the point where the cost per correct response of a designed agent becomes lower than the agents in the initial agents library, we combine the inference cost of the best agent (Figure \ref{fig:Figure_BestAgentCost}) with the fixed cost of agent design. The fixed design cost, $C_{\text{0}}$, includes the total cost of all the sampling step (Algorithm \ref{alg:meta-agent}, line $6$; Figure \ref{fig:fixed_costs}) and evaluation costs to compute $s_i$ (Algorithm \ref{alg:meta-agent}, line $7$). The total cost of an agent is the sum of $C_{\text{0}}$ and a per-example inference cost, $C_{\text{j}}$:\(
C_{\text{0}} + n \cdot C_{\text{j}}
\). 

\begin{figure}[h]
  \centering
\includegraphics[width=0.48\textwidth]{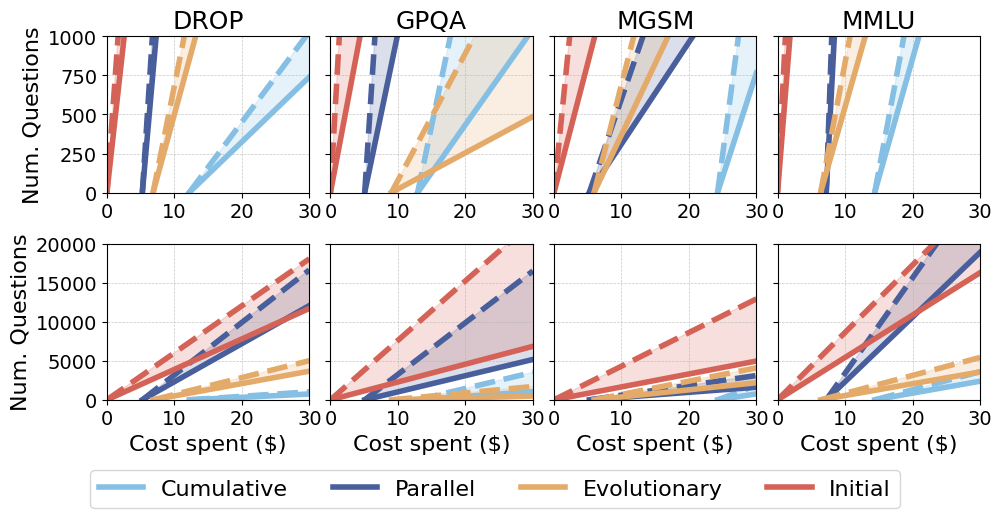}
  \caption{\textbf{Cost Efficiency: Highest performing agent from the initial library generates the outputs with same total performance at lower cost.}
  Number of questions solved (solid lines) and attempted (dashed lines) versus cost spent for agents with best training set performance. The x-intercept indicates the fixed cost $C_0$ ($0$ for agents in initial library); the slope beyond reflects variable cost per attempt or per solution.}
  \label{fig:fig3_2}
\end{figure}

In Figure \ref{fig:fig3_2}, the intersection of the red solid line with another solid line marks the break-even point, where deploying the meta-agent lowers the cost per correct response. This occurs at approximately $n=15{,}000$ examples for DROP and MMLU with parallel context curation. In contrast, for other datasets and context curation methods, performance gains do not justify the associated costs at any scale.
\section{Conclusion}
Our analysis highlights key trade-offs between (1) final performance and behavioral diversity and (2) performance relative to cost. Evolutionary context curation boosts performance but reduces diversity. While meta-agent-driven design can produce cost-effective agents in some cases, the performance gains rarely justify the increased design and inference costs, even at scale.

\newpage
\section*{Limitations}
\paragraph{Scope} Our work focuses on a class of meta-agent approaches that follow the sample–evaluate–iterate pattern. While restricting our scope to this setup enables us to highlight general patterns, our findings may not apply directly to the broader space of possible meta-agent paradigms.

\paragraph{Evaluation} We evaluate performance primarily in terms of accuracy and F-1 scores. Our findings may not directly translate to domains where consistency is critical, or where different utility metrics are more appropriate. 

\paragraph{Economic Viability} Our analysis of economic viability is most suited for domains with strong verifiers as it emphasizes the cost of sampling a correct or high-performing answer. Other formulations may be better suited for different applications. 

\paragraph{Similarity Computation} Cosine similarity favors alignment between agents that succeed on the same examples. The metric reaches its maximum ($1$) when agents can solve the same set of questions. However, favoring alignment introduces an overall bias toward high-performing agents. Due to this bias, high-performing agents appear more similar, whereas agents that fail consistently appear orthogonal. As a robustness check, we also computed Hamming distances between binary score vectors and observed similar trends (Figure~\ref{fig:hamming}, \ref{fig:hamming2}).

\paragraph{Meta Evaluation} Meta-agent evaluations involve multiple sources of stochasticity, including (1) LM output randomness, (2) error propagation in chained reasoning inside agents, (3) sampling variability of the meta-agent, (4) stochasticity in evaluation results for the designed agents, which can then lead the trajectories in different directions, and (5) meta trajectory-level divergence due to the differences in chains of sampled agents and their evaluation scores. Robust evaluation thus requires multiple trajectory samples for reliable conclusions. Due to the extensive costs of larger scale evaluations, the results we present are averaged across 3 runs.

\section*{Safety Considerations} 
For meta-agents, the unchecked generation and execution of complex systems may present safety risks. Such systems are difficult to audit or control prior to deployment within automated design loops.

\section*{Acknowledgments}
We would like to thank Aakanksha Chowdhery and Azalia Mirhoseini for helpful discussions and feedback. Batu El gratefully acknowledges the support of the Knight-Hennessy Scholarship. We acknowledge the use of AI tools to assist with language refinement during the writing process and code development.

\bibliography{custom}

\appendix
\section{Experimental Setup Details}
\subsection{Initial Agents Library}
\label{apdx:initial-agents}
Our initial agent library, \( F \), consists of the following methods:  
(1) \textbf{Chain-of-Thought} \citep{wei2023chainofthoughtpromptingelicitsreasoning}, which prompts the language model to output its reasoning before arriving at an answer;  
(2) \textbf{Majority Voting}, which selects the consensus response from multiple generated answers;  
(3) \textbf{Refinement from Feedback} \citep{madaan2023selfrefineiterativerefinementselffeedback}, where the model iteratively improves its answer based on self-feedback;  
(4) \textbf{LLM-Debate} \citep{du2023improvingfactualityreasoninglanguage}, where multiple language model instances are prompted to debate with each other;  
(5) \textbf{Quality-Diversity} \citep{lu2024intelligentgoexplorestandingshoulders}, which generates and ensembles diverse responses;  
(6) \textbf{Routing}, which directs tasks to the most appropriate language model instances prompted to behave like an expert of a subject; and 
(7) \textbf{Stepping-back} \citep{hu2024automateddesignagenticsystems}, which encourages the model to first reflect on relevant scientific principles before answering. This is consistent with the setup in \citet{hu2024automateddesignagenticsystems}.

\subsection{Experimental Setup}
\label{apdx:exp-setup}

\textbf{Number of Iterations.} In all our experiments, we use $T = 30$. 

\paragraph{Dataset Size.} For each of our MGSM, MMLU, DROP datasets, we select \( 128 \) examples from our dataset as training examples, denoted as \( D_{\text{train}} \), and \( 200 \) examples as test examples, denoted as \( D_{\text{test}} \). For GPQA, we use $32$ samples as training examples and the remaining $160$ samples as test examples. To reduce the variance during training, we use each training example from GPQA $5$ times and compute scores using $5 \times 32 = 160$ evaluations. Performance is measured using F1-score for DROP and accuracy for the other datasets.

\paragraph{Models.} In our experiments, we use \texttt{gpt-3.5} as the engine of the \texttt{LanguageModel} class. We use a larger, more powerful model, \texttt{gpt-4o},  as the engine of the meta-agent. This is consistent with the setup in \citet{hu2024automateddesignagenticsystems}.

\section{Other Related Works}
\paragraph{Agentic Systems}
Agentic systems have demonstrated remarkable success across a range of domains. Several agentic systems have advanced scientific automation, including frameworks for end-to-end research \citep{lu2024aiscientistfullyautomated}, autonomous paper writing \citep{yamada2025aiscientistv2workshoplevelautomated}, nanobody design in a virtual lab \citep{Swanson2024}, and multi-agent ideation \citep{su2025headsbetteroneimproved}. Beyond research, agentic systems have demonstrated effectiveness in complex operational contexts, including generalist problem-solving \citep{fourney2024magenticonegeneralistmultiagentsolving, lu2025octotoolsagenticframeworkextensible}, financial modeling and trading \citep{okpala2025agenticaisystemsapplied,xiao2025tradingagentsmultiagentsllmfinancial}, and robotics manipulation \citep{singh2024malmmmultiagentlargelanguage}.

\paragraph{Recursive Self-Improvement}
STOP \citep{zelikman2024selftaughtoptimizerstoprecursively}, Promptbreeder \citep{fernando2023promptbreederselfreferentialselfimprovementprompt}, Gödel Agent \citep{yin2025godelagentselfreferentialagent}, and \citet{zhou2024symboliclearningenablesselfevolving} implement recursive self-improvement by enabling agents to iteratively refine their own prompts, code, or internal reasoning logic.

\section{Additional Results}
\begin{table}[h]
\centering
\resizebox{0.45\textwidth}{!}{%
\begin{tabular}{lccccc}
\toprule
Setting & DROP & MGSM & MMLU  & GPQA  & Avg. \\
\midrule
C & 96.6 & 89.1 & 99.2 & 91.9 & 94.2 \\
P & 96.0 & 95.3 & 97.7 & 94.4 & \cellcolor{green!20}\textbf{95.9} \\
E & 93.6 & 93.0 & 99.2 & 91.9 & 94.4 \\
\bottomrule
\end{tabular}%
}
\caption{Coverage. Proportion of questions correctly answered at least by one of the designed agents. The designed agents includes all 90 agents designed across 3 runs.}
\label{tab:coverage}
\end{table}

\begin{figure}[h]
  \centering
  \includegraphics[width=0.45\textwidth]{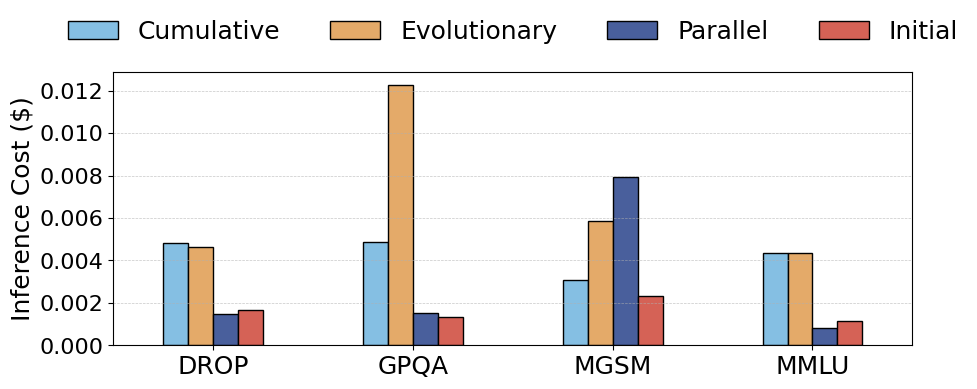}
  \caption{\textbf{Average inference cost per test query of the best agents.} For best agent in the initial library $F$ (Initial, see Appendix \ref{apdx:initial-agents}), best agent designed by meta agent with \(\phi_C\) (Cumulative), best agents designed by meta agent with \(\phi_P\) (Parallel) , best agent designed by meta agent with \(\phi_E\) (Evolutionary). Averaged across the single best agents from 3 runs. Best agent is selected based on the highest training performance.}
\label{fig:Figure_BestAgentCost}
\end{figure}

\begin{figure*}[t]
  \centering
  \includegraphics[width=\textwidth]{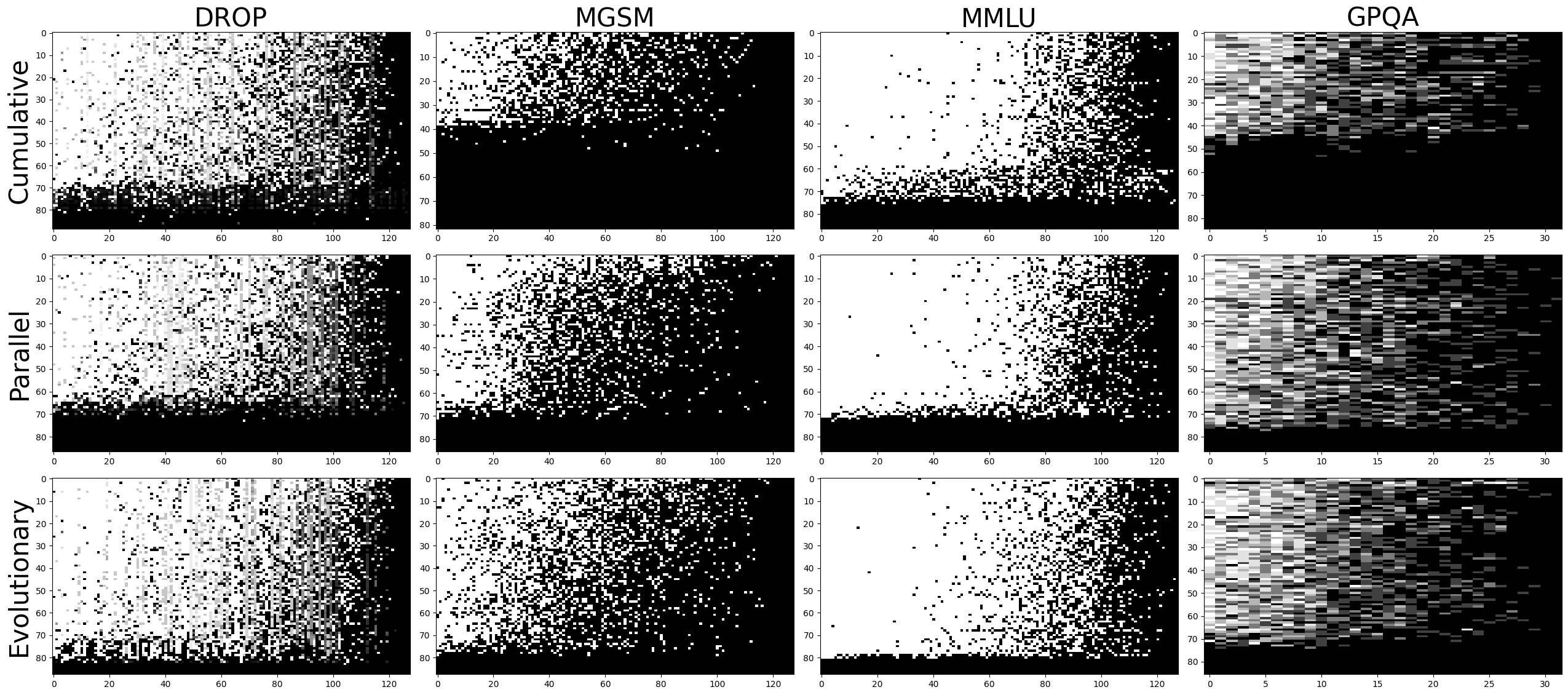}
\caption{Score matrix $\mathbf{S}$, where each row corresponds to an agent and each column to a dataset example. A cell is white if the agent answers correctly and black otherwise. For DROP, gray indicates intermediate F1 scores; for GPQA, gray denotes partial correctness across repeated attempts. The normalized rows, $s_i$, serve as agent embeddings, capturing performance across training questions.}
\label{fig:S}
\end{figure*}

\begin{figure*}[t]
  \centering
  \includegraphics[width=\textwidth]{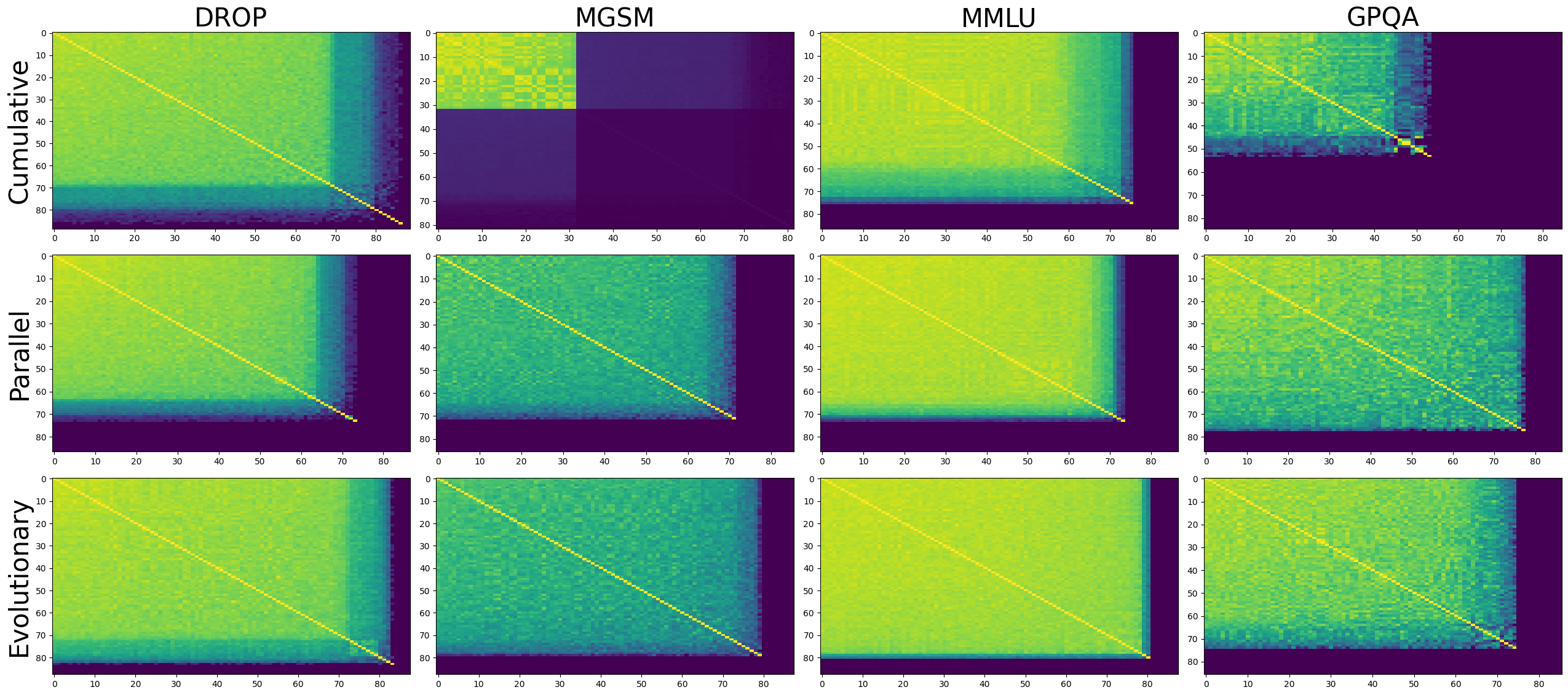}
  \caption{Cosine similarity matrix $\mathbf{C}$, with agents reordered by descending average similarity to all other agents.}
  \label{fig:SST}
\end{figure*}

\begin{figure*}[t]
  \centering
\includegraphics[width=\textwidth]{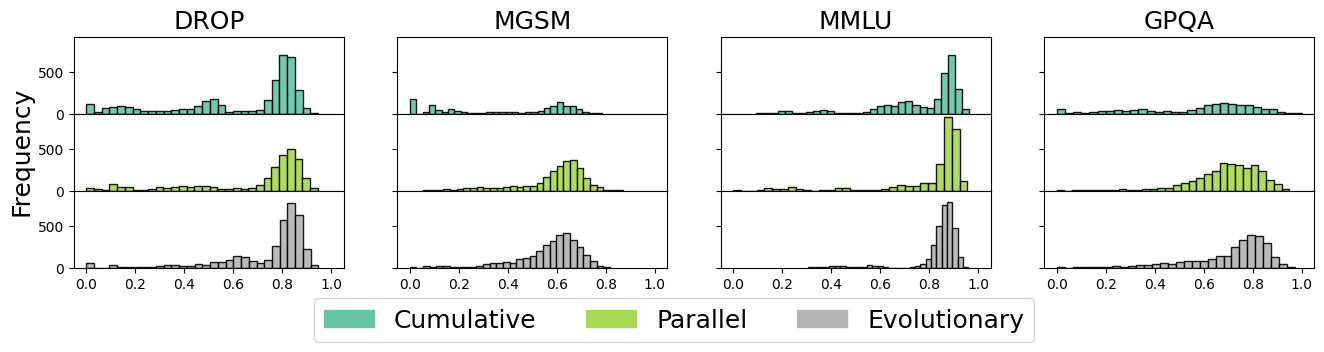}
  \caption{Histograms of agent similarities (entries of $\mathbf{C}$), excluding agents with zero performance (all black rows of $\mathbf{S}$ in Figure \ref{fig:S}, and corresponding dark blue rows and columns of $\mathbf{C}$ in Figure \ref{fig:SST}). Only the upper triangular entries of $\mathbf{C}$ (excluding the diagonal) are used, as $\mathbf{C}$ is symmetric. Each subplot shows histograms of similarity scores (x-axis) and their frequency (y-axis).}
  \label{fig:Figure6_entries}
\end{figure*}

\begin{figure*}[t]
  \centering
\includegraphics[width=\textwidth]{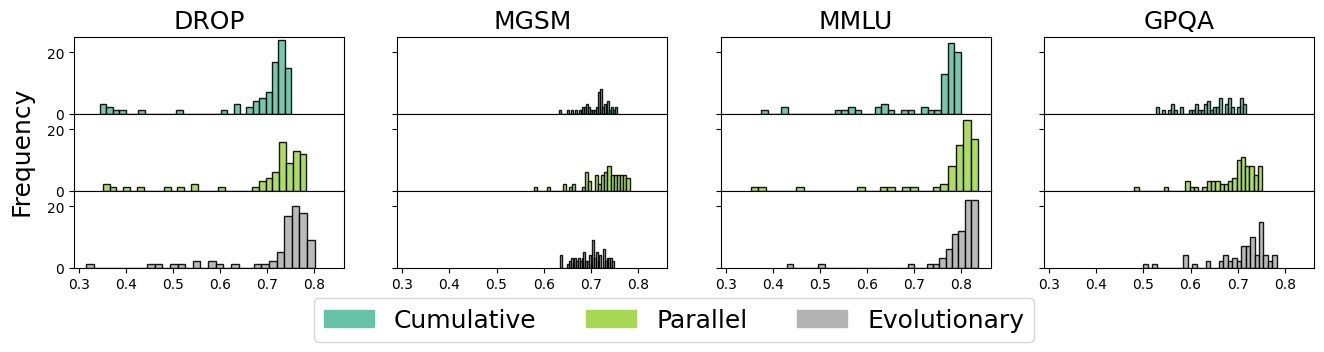}
  \caption{Figure \ref{fig:Figure6_rowavg} with (1 - Hamming distance) as the similarity metric. All nonzero entries of $S$ are set to $1$.}
  \label{fig:hamming}
\end{figure*}

\begin{figure*}[t]
  \centering
\includegraphics[width=\textwidth]{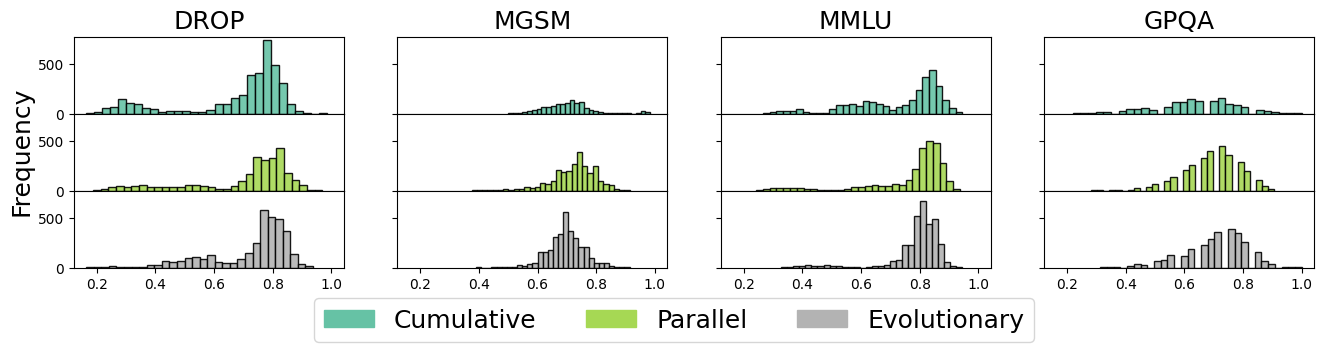}
  \caption{Figure \ref{fig:Figure6_entries} with (1 - Hamming distance) as the similarity metric. All nonzero entries of $S$ are set to $1$.}
  \label{fig:hamming2}
\end{figure*}

\begin{figure*}[t]
  \centering
\includegraphics[width=\textwidth]{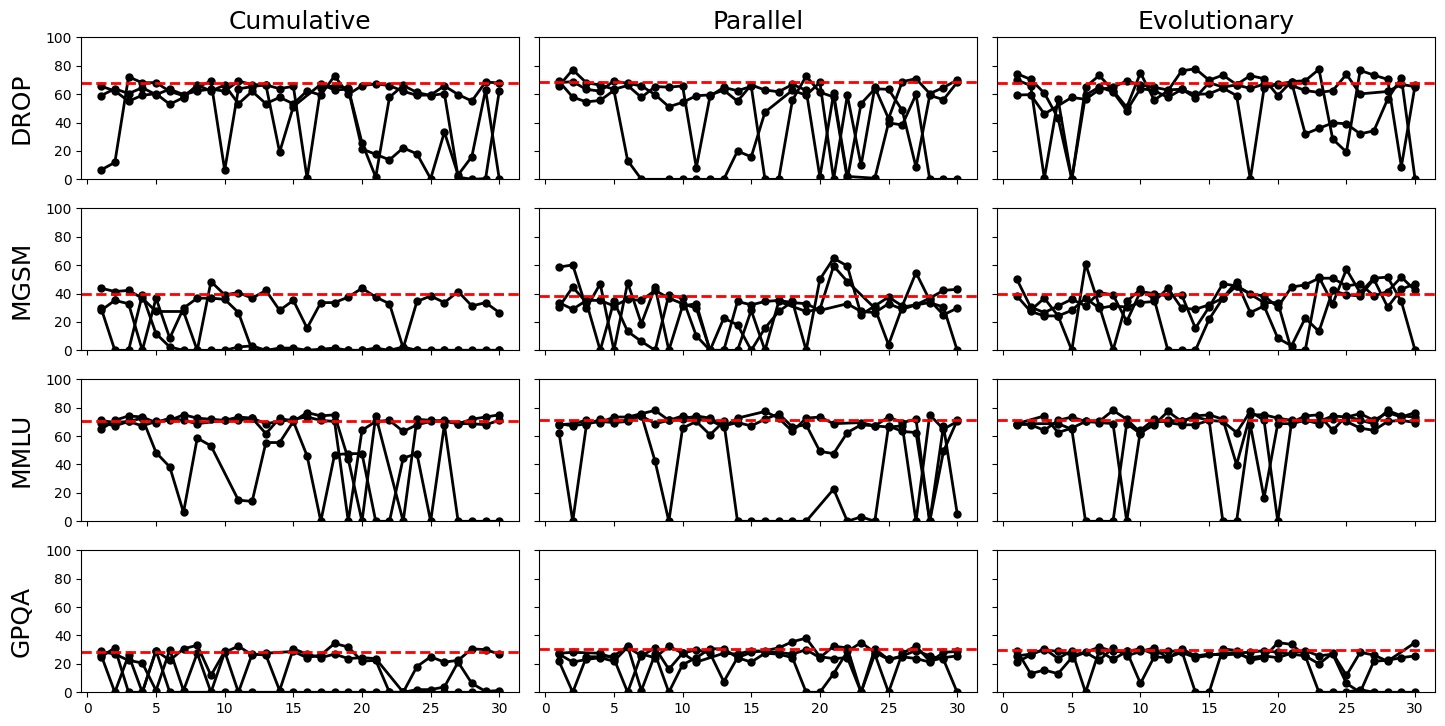}
  \caption{Training performance of designed agents across iterations. The dotted red line shows the performance of the best agent from the initial library.}
  \label{fig:}
\end{figure*}

\begin{figure*}[t]
  \centering
  \includegraphics[width=\textwidth]{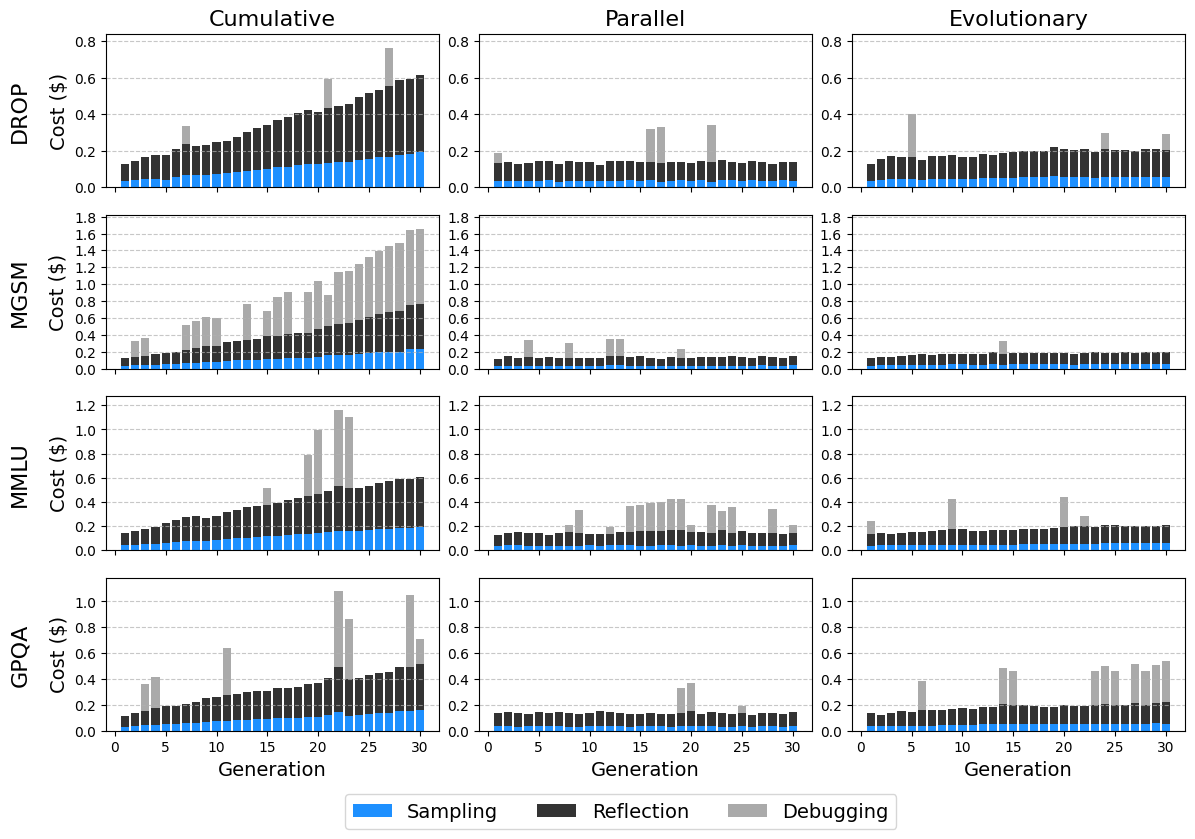}
\caption{Design cost of the next agent across iterations. While costs remain stable with Parallel and Evolutionary context curation, they increase linearly with increasing context length in Cumulative context curation.}
  \label{fig:fixed_costs}
\end{figure*}

\end{document}